\PassOptionsToPackage{table}{xcolor}
\documentclass[preprint,12pt,authoryear]{elsarticle}

\usepackage{amssymb}
\usepackage{amsmath}

\usepackage{algorithmic}
\usepackage{algorithm}
\usepackage{array}
\usepackage{makecell}
\usepackage[caption=false,font=normalsize,labelfont=sf,textfont=sf]{subfig}
\usepackage{textcomp}
\usepackage{stfloats}
\usepackage{orcidlink}
\usepackage{url}
\usepackage{verbatim}
\usepackage{graphicx}
\usepackage{cite}
\usepackage{booktabs}
\usepackage{multirow}
\usepackage{microtype}
\usepackage{adjustbox}
\usepackage{float}
\usepackage{pifont}
\global\long\def\newmodelshort{\textrm{{GeoVis-GNN}}}
\global\long\def\newdatasetshort{\textrm{{MPHOI-120}}}
\definecolor{Cerulean}{rgb}{0.0, 0.48, 0.65}

\usepackage{xspace}
\usepackage{xcolor}

\journal{Expert Systems with Applications}

\makeatletter
\DeclareRobustCommand\onedot{\futurelet\@let@token\@onedot}
\def\@onedot{\ifx\@let@token.\else.\null\fi\xspace}

\def\eg{\emph{e.g}\onedot}

\makeatother

\begin{document}

\begin{frontmatter}

\title{Geometric Visual Fusion Graph Neural Networks for Multi-Person Human-Object Interaction Recognition in Videos} %

\author[1]{Tanqiu Qiao}
\author[1]{Ruochen Li}
\author[1]{Frederick W. B. Li}
\author[2]{Yoshiki Kubotani}
\author[3]{Shigeo Morishima}
\author[1]{Hubert P. H. Shum \corref{4}}

\cortext[4]{Corresponding author: Hubert P. H. Shum, Department of Computer Science, Durham University, Durham, DH1 3LE, United Kingdom. \\
\hspace*{2em}E-mail addresses: tanqiu.qiao@durham.ac.uk (T. Qiao), ruochen.li@durham.ac.uk (R. Li), frederick.li@durham.ac.uk (F. W. B. Li), yoshikikubotani.lab@gmail.com (Y. Kubotani), shigeo@waseda.jp (S. Morishima), hubert.shum@durham.ac.uk (H. P. H. Shum).}

\affiliation[1]{organization={Durham University},%
            addressline={Department of Computer Science}, 
            city={Durham},
            postcode={DH1 3LE}, 
            country={UK}}
\affiliation[2]{organization={cvpaper.challenge},
            city={Tokyo},
            country={Japan}}
\affiliation[3]{organization={Waseda University},
            addressline={Waseda Research Institute for Science and Engineering},
            city={Tokyo},
            postcode={169-8555},
            country={Japan}}

\begin{abstract}
  Human-Object Interaction (HOI) recognition in videos requires understanding both visual patterns and geometric relationships as they evolve over time. Visual and geometric features offer complementary strengths. Visual features capture appearance context, while geometric features provide structural patterns. Effectively fusing these multimodal features without compromising their unique characteristics remains challenging. We observe that establishing robust, entity-specific representations before modeling interactions helps preserve the strengths of each modality. Therefore, we hypothesize that a bottom-up approach is crucial for effective multimodal fusion. Following this insight, we propose the Geometric Visual Fusion Graph Neural Network (GeoVis-GNN), which uses dual-attention feature fusion combined with interdependent entity graph learning. It progressively builds from entity-specific representations toward high-level interaction understanding. To advance HOI recognition to real-world scenarios, we introduce the Concurrent Partial Interaction Dataset (MPHOI-120). It captures dynamic multi-person interactions involving concurrent actions and partial engagement. This dataset helps address challenges like complex human-object dynamics and mutual occlusions. Extensive experiments demonstrate the effectiveness of our method across various HOI scenarios. These scenarios include two-person interactions, single-person activities, bimanual manipulations, and complex concurrent partial interactions. Our method achieves state-of-the-art performance.
\end{abstract}

\begin{keyword}
Human-object interaction \sep 
Multi-person interaction \sep 
Feature fusion

\end{keyword}

\end{frontmatter}

\section{Introduction}
Human-Object Interaction (HOI) recognition aims to interpret the intricate relationships between humans and the objects they interact with. While traditional video analysis tasks can achieve strong performance using visual features alone, HOI recognition demands additional geometric reasoning through human poses and object spatial configurations. In video-based scenarios, this complexity intensifies as systems track dynamic spatial relationships across frames while handling occlusions and viewpoint changes. This complexity goes beyond the pixel-level understanding required for coarse actions like cooking. Instead, it includes geometric analysis of fine-grained interactions, such as specific hand poses for holding objects or spatial configurations needed for cutting. These interactions often occur concurrently or in sequence.

Significant efforts have focused on image-based HOI detection, which combines object localization and interaction classification within static frames. Recent advances have leveraged transformer architectures with specialized mechanisms \citep{ma2023fgahoi,kim2023relational,li2024neural,zhu2024geometric}. While these methods are effective for static scenes, they are inadequate for capturing the temporal dynamics and motion complexities inherent in video scenarios.
Video-based HOI recognition is a relatively less-explored area, which requires understanding not only the spatial relationships between humans and objects but also how these interactions evolve over time. Existing methods primarily rely on visual features \citep{morais2021learning,tu2022video,wang2023exploring}, which encode rich appearance and contextual cues but are vulnerable to occlusions common in real-world scenarios. In contrast, geometric features, derived from human pose estimations and object spatial configurations, provide explicit structural details crucial for interaction understanding \citep{wan2019pose,das2020vpn,zhu2024geometric}. While some recent approaches \citep{qiao2022geometric,qiao2024category} have attempted to integrate visual and geometric features to leverage their complementary advantages, these methods typically fuse both modalities directly. This simplistic fusion neglects the distinct and valuable characteristics inherent to each modality, resulting in suboptimal interaction representations. Consequently, effectively integrating visual and geometric features remains challenging, limiting the capability to fully exploit their complementary strengths for robust interaction recognition.

Another challenge is that current approaches often fail to preserve fine-grained entity characteristics when integrating multimodal features, resulting in the loss of important interaction details. Effectively fusing geometric and visual features to fully leverage their potential for enhancing entity interaction recognition remains underexplored. A mixed-fusion approach \citep{qiao2022geometric} that combines feature fusion and entity interaction learning in a unified graph suffers from entangled entity representations, limiting explicit HOI understanding. A top-down approach \citep{zheng2023deep} that prioritizes entity-level relationships over specific features may miss critical interaction details and misalign overarching patterns with individual nuances. An alternative bottom-up approach \citep{wang2022regularizing,wang2022lite} starts with fundamental entity features, enabling detailed integration before addressing complex interactions, resulting in more effective entity interaction analysis. Although the bottom-up approach has potential benefits, it faces challenges in accurately fusing multimodal features. Specifically, it struggles to preserve fine-grained entity characteristics when transitioning from individual entity representations to modeling complex interactions.

In this paper, we introduce a novel Geometric Visual Fusion Graph Neural Network ($\newmodelshort$) to address the two critical challenges identified in previous research. To overcome the inadequate fusion of geometric and visual features, we propose a dual-attention mechanism operating at the feature level. This mechanism first utilizes graph attention to produce structured geometric embeddings and then employs channel attention to adaptively fuse these embeddings with visual features, effectively leveraging the complementary strengths of both modalities. To preserve fine-grained entity characteristics, we adopt a bottom-up approach. Specifically, we first establish robust entity-specific representations and then progressively build up to higher-level interaction understanding through an interdependent entity graph. This graph models explicit spatial interactions and implicit contextual dependencies among entities, ensuring that detailed entity characteristics are retained even when modeling complex interactions.

\begin{figure}[t]
\centering
\includegraphics[width=\linewidth]{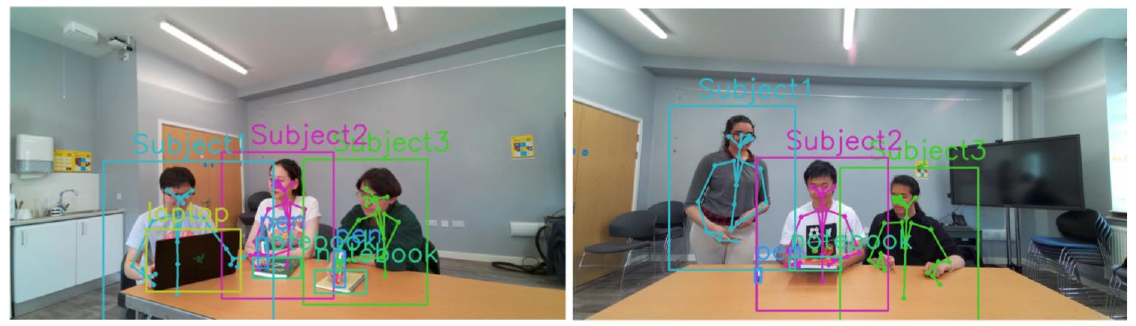}
\caption{Two examples (\textit{Teaching} and \textit{Signing}) of our collected concurrent partial interaction datasets. Geometric features such as skeletons and bounding boxes are annotated.}
\label{fig:geometric_feature}
\end{figure}

Video-based HOI datasets have primarily focused on single-person activities or limited two-person interactions, leaving a notable gap in capturing the complexity of real-world multi-person scenarios. Existing datasets, such as UCLA HHOI \citep{shu2016learning,shu2017learning} and MPHOI-72 \citep{qiao2022geometric}, include interactions with up to two participants and a few objects. They assume all participants are continuously active, which limits their ability to represent scenarios with idle or waiting individuals. To address this limitation, we introduce the Concurrent Partial Interaction Dataset ($\newdatasetshort$), which captures dynamic multi-person interactions where some participants are engaged while others are idle (Fig.~\ref{fig:geometric_feature}). This dataset incorporates diverse interactions, high variability, and challenging dynamics such as simultaneous actions, mutual dependencies, and occlusions. It offers a richer and more realistic benchmark for advancing HOI recognition in complex real-world scenarios.

We demonstrate the effectiveness of our approach across a comprehensive spectrum of real-world HOI scenarios. These include single-person interactions in CAD-120 \citep{koppula2013learning}, bimanual manipulations in Bimanual Actions \citep{dreher2020learning}, two-person collaborative activities in MPHOI-72 \citep{qiao2022geometric}, and concurrent partial interactions in our proposed $\newdatasetshort$ dataset. These diverse datasets collectively represent the full range of human-object interactions, from individual tasks to simultaneous multi-person collaborations. Our main contributions are:

\begin{itemize}
  \item A novel bottom-up framework ($\newmodelshort$) for multi-person HOI recognition. It preserves fine-grained entity characteristics and progressively builds from entity-level representations to interaction-level understanding.~\footnote{Dataset and code are available at \url{https://github.com/tanqiu98/GeoVis-GNN}.}  
  
  \item A dual-attention fusion mechanism that first employs graph attention to learn structured geometric embeddings, followed by channel attention-based adaptive fusion with visual features, resulting in rich entity-specific representations.

  \item An interdependent entity graph that leverages the enriched entity representations to simultaneously model explicit spatial interactions and implicit contextual dependencies among multiple entities.

  \item A novel Concurrent Partial Interaction Dataset ($\newdatasetshort$)\footnote[2]{Data collection performed in the UK, under Durham University Ethics Approval Ref: \texttt{COMP-2020-10‐01T19\_29\_22‐cbmw62}.} captures complex multi-person interactions with concurrent actions and partial engagement, providing a more realistic benchmark to advance HOI recognition.
\end{itemize}

\section{Related Work}
\label{sec:relatedwork}
\subsection{HOI Recognition}
HOI recognition is divided into two primary areas: HOI detection in images and HOI recognition in videos. HOI detection in images focuses on identifying interactions within a single static picture, combining object localization with interaction classification. It aims to detect triplets $\langle human, verb, object \rangle$, providing a spatially grounded understanding of HOIs in a single image \citep{kim2023relational,cheng2024parallel,li2024neural}. These methods are not directly applicable to HOI recognition in videos, as the task introduces a temporal dimension, requiring models to capture interactions as they evolve over time. Video-based HOI recognition demands the ability to process dynamic, sequential data to understand interaction context more comprehensively. While some image-based methods provide valuable spatial insights using geometric and visual features \citep{wu2022mining,park2023viplo,zhu2024geometric}, they lack the temporal modeling capabilities necessary to capture motion patterns, action progression, and continuity. This limitation results in an incomplete understanding of the evolving interactions critical for accurate recognition in video scenarios.

HOI recognition in videos encompasses human action analysis \citep{tan2022bidirectional,hu2024human,cob2024new} and skeleton-based activity recognition \citep{setiawan2022sequential,cheng2024dense,yu2024exploring} by integrating the detection of human movements and postures with the contextual understanding of interactions between humans and objects, thereby offering a more holistic approach to activity recognition in complex environments. Deep Neural Networks (DNNs) and graphical models are combined in recent works. \citet{wang2021spatio} utilize the parsed graphs to directly model the global relationship between the human and object, capturing the state change of the interacting objects across frames. ASSIGN \citep{morais2021learning} presents a visual feature attention model to learn asynchronous and sparse HOIs in videos. TUTOR \citep{tu2022video} employs a reinforced tokenization strategy that jointly learns instance tokens through selective attention and aggregation in the spatial domain and links them across frames to generate tubelet tokens, serving as highly-abstracted spatio-temporal representations for HOI recognition. \citet{xing2022understanding} introduce a spatial attention mechanism that can enhance action recognition by adaptively generating a spatial-relation graph during HOIs. STIGPN \citep{wang2023exploring} exploits spatio-temporal graph convolutions to enhance the detection of salient human-object interactions and efficiently modeling long-term dynamics. 

Based on prior visual-based approaches, 2G-GCN \citep{qiao2022geometric} firstly proposes the multi-person HOI recognition problem and incorporates geometric features into the Graph Convolutional Network (GCN). However, 2G-GCN merges the collective geometric features of all entities with individual visual features, leading to potential hierarchical misalignment. The high-level spatial information from geometric features may not align well with detailed, entity-specific visual data. As a result, the model may struggle to correctly distinguish between different entities and their interactions, leading to impaired performance and a focus on less relevant objects. CATS \citep{qiao2024category} learns HOIs from multimodal feature fusion of different categories, such as humans and objects, to the scenery interactive graph. However, it neglects the entity concept and entity relationships within the same category, which is particularly limiting in multi-person HOI scenarios. Therefore, in this paper, we follow a bottom-up approach that first establishes fine-grained entity-specific features before capturing entity-level interactions, ensuring precise entity representations and facilitating accurate interaction modeling in complex multi-entity scenarios.

\subsection{Geo-Vis Fusion in Human Activities}
Combining diverse data modalities offers unique, complementary insights that lead to a more holistic understanding of a subject. In multimodal research of human action recognition, attention has been directed towards key areas of the human body, particularly the hands \citep{baradel2017human,Baradel2018human,baradel2018glimpse}. These studies employ attention-based methods to improve the overall accuracy of models that integrate skeletal and visual modalities. Building on this, \citet{bruce2021multimodal} expand the focus to include additional regions of the body such as the head, hands, and feet by adopting a temporal approach. They generate a fused representation by multiplying spatial attention weights with appearance features. TSMF \citep{bruce2021multimodal} fuses skeleton and RGB data at the model level using teacher-student networks to learn enriched representations. However, these model-based fusion models often lack transparency, making it difficult to interpret how individual features contribute to recognition. Besides, \citet{boulahia2021early} investigate the integration of various image modalities (RGB, Depth, Skeleton, and InfraRed) at different stages of the action recognition pipeline, encompassing early, intermediate, and late fusion techniques, to enhance the robustness of recognition. 

In human interaction analysis, \citet{wan2019pose} concatenate human skeletal embeddings with visual embeddings from other branches like human, object and union to obtain the final holistic feature in the HOI scene. \citet{zhou2022human} combine embedded visual and human pose features through element-wise addition. \citet{wang2023exploring} directly concatenate multimodalities to output visual-spatial and spatial-semantic feature sequences, which are then input into a two-stream network. CATS \citep{qiao2024category} also concatenates geometric and visual features for different categories. However, their direct operations may dilute distinct contributions of visual and geometric features, often amplifying dominant features while undervaluing subtle geometric cues, which can reduce accuracy in fine-grained interaction recognition. These challenges indicate that the fusion of geometric and visual features still has design intricacies that require further optimization. Therefore, we propose a dual-attention fusion mechanism to integrate geometric and visual features. This mechanism combines graph attention and channel attention to preserve the complementary strengths of both modalities. As a result, it produces enriched entity representations, enabling more robust and precise interaction modeling.

\subsection{Video-Based HOI Datasets}
There are various datasets available for the investigation of HOI in videos for multiple tasks. For single-person HOI recognition, datasets like CAD-120 \citep{koppula2013learning}, Bimanual Actions \citep{dreher2020learning}, Bimanual Manipulation \citep{KrebsMeixner2021}, \textit{etc.} are effective, with the latter two also encompassing bimanual HOI tasks due to their focus on interactions involving both hands. There are several datasets available for single-hand HOI recognition tasks, including Something-Else \citep{materzynska2020something}, VLOG \citep{fouhey2018lifestyle}, EPIC Kitchens \citep{Damen2021RESCALING}. Since EPIC Kitchens records both hands in the cooking process, it can also be utilized for bimanual HOI recognition. Besides, a full-body HOI dataset called BEHAVE \citep{bhatnagar2022behave} includes multi-view RGBD frames, associated 3D SMPL and object fits. HOI4D \citep{liu2022hoi4d} is a large-scale 4D egocentric dataset aimed to facilitate research on category-level HOIs. The UCLA HHOI Dataset \citep{shu2016learning,shu2017learning} focuses on human-human-object interaction with a maximum of two humans and one object involved. The MPHOI-72 dataset \citep{qiao2022geometric} is specifically proposed for the multi-person HOI recognition task but is constrained to interactions between two individuals and 2-4 objects, reducing its applicability to complex real-world scenarios.

\section{Concurrent Partial Interaction Dataset}
The majority of video-based HOI datasets primarily focus on single-person HOIs, albeit from various perspectives \citep{koppula2013learning,Damen2021RESCALING,bhatnagar2022behave,liu2022hoi4d}. Efforts to encompass multiple human interactions are still in their infancy. For instance, the UCLA HHOI dataset \citep{shu2016learning,shu2017learning} captures interactions involving up to two people and one object, while MPHOI-72 \citep{qiao2022geometric} slightly broadens this scope to include two people and several objects. However, these datasets assume that all participants are continuously active throughout the activity. In contrast, real-world multi-person HOIs often include scenarios where some individuals are not interacting, such as sitting or standing idle while waiting for their turn. This gap highlights the need for datasets that better represent the complexity and variability of real-world multi-person and multi-object interactions.

To bridge this critical gap, we introduce the Concurrent Partial Interaction Dataset ($\newdatasetshort$), which captures dynamic interactions involving multiple people and objects. In our context, “concurrent" refers to scenarios where multiple interactions occur simultaneously, while “partial interaction" highlights moments when not all individuals are actively engaging - some may remain idle or waiting during certain moments of the activity. For example, in the \textit{Signing} activity, while two people are passing a notebook and pen, the other person is standing or sitting idle. Similarly, when one individual is signing, the other two are not interacting. Such scenarios, which reflect real-world interaction patterns, are extensively captured in our dataset, providing a richer benchmark for advancing multi-person HOI recognition methods. In addition, increasing the number of people and objects introduces an exponential increase in complexity. It expands the range of human-human, human-object, and object-object interactions, while also intensifying challenges such as simultaneous actions, mutual dependencies, and significant occlusions.

\begin{figure}
\centering
\includegraphics[width=\linewidth]{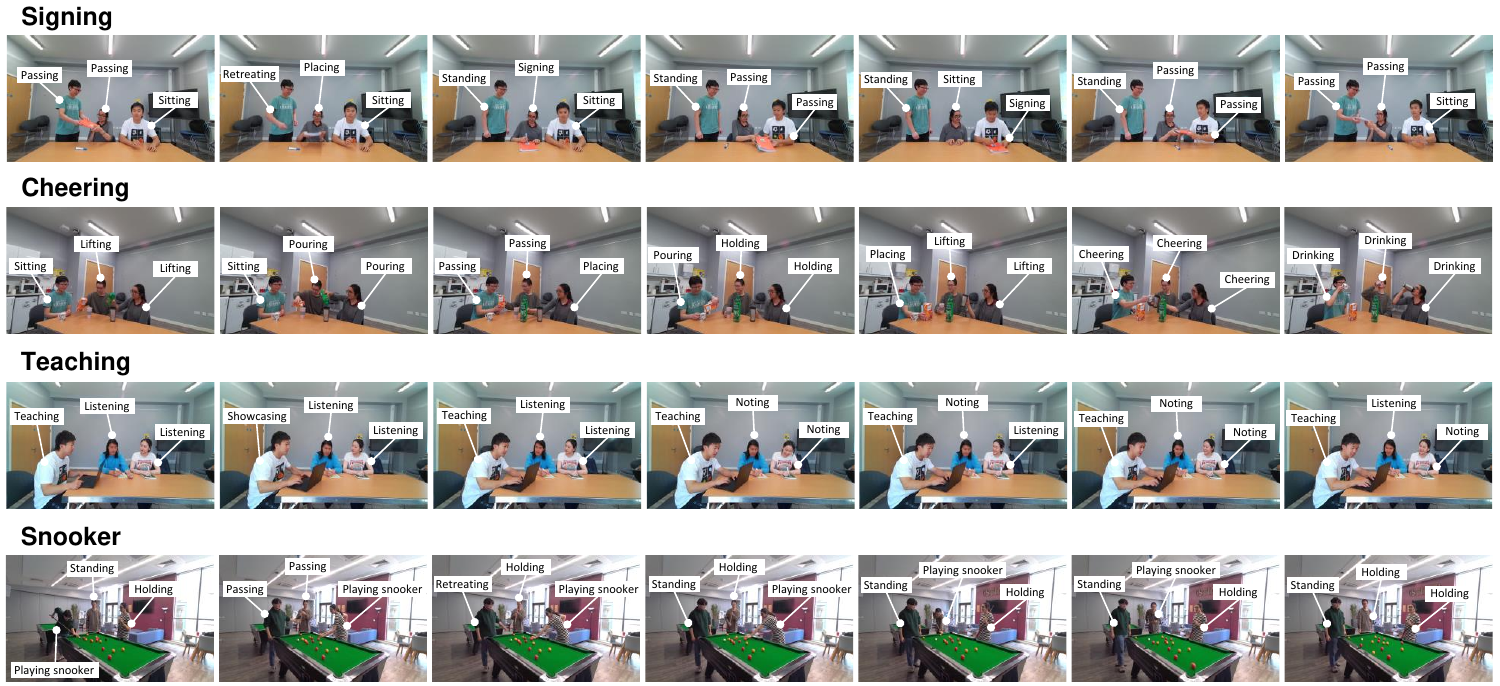}
\caption{Sample video screenshots from our new $\newdatasetshort$ dataset, displaying concurrent partial interactions along the timelines of four multi-person HOI activities in daily life.}
\label{fig:mphoi120_dataset}
\end{figure}

\subsection{Dataset Details}
$\newdatasetshort$ is a dataset of 120 high-resolution videos of three participants interacting with 2 to 5 objects. All annotations are performed frame-by-frame by a single trained annotator using a predefined list of sub-activities to ensure consistency and avoid inter-annotator variability. Sample video screenshots with annotated sub-activities for all activities are shown in Fig.~\ref{fig:mphoi120_dataset}. Each main activity captures unique interaction patterns. \textit{Signing} highlights turn-taking behaviors amid potential occlusions, while \textit{Cheering} features synchronous and sequential human-object actions. \textit{Teaching} depicts fine-grained states (e.g., noting vs. listening) between a teacher and students, and \textit{Snooker} focuses on strategic turn-taking with frequent body occlusions around the table.

Leveraging the Azure Kinect SDK along with the Body Tracking SDK \citep{Kinect2022}, we acquire RGB-D videos to capture the comprehensive dynamics of multiple individual skeletons. We offer 2D human skeletal data and bounding boxes for both subjects and objects within each video, serving as geometric characteristics. The integration of depth information within our dataset further broadens its utility, such as versatile benchmarks for 3D human pose estimation \citep{you2023co,Zhai_2023_ICCV} and 3D object estimation \citep{fan2023occ,heitzinger2023fast}, among others.

\begin{table}[h!]
\caption{A statistical comparison between $\newdatasetshort$ and popular HOI datasets. CPI denotes Concurrent Partial Interactions.}
\label{tab:dataset_comparison}
\centering
\begin{adjustbox}{width=\linewidth}
\begin{tabular}{lcccc}
\toprule
Datasets & $\newdatasetshort$ & MPHOI-72 & CAD-120 & Bimanual Actions\tabularnewline
\midrule
No. people interacting & 3 & 2 & 1 & 1 \tabularnewline
Total videos & 120 & 72 & 120 & 540 \tabularnewline
Total frames & 53604 & 26383 & 61585 &  221000 \tabularnewline
Total frames of CPI & 20100 & 0 & 0 & 0 \tabularnewline
Video average length & 15s & 12s & 17s & 15s \tabularnewline
No. sub-activities & 17 & 13 & 10 & 14 \tabularnewline
No. subjects/objects & 7/6 & 5/6 & 4/10 & 6/12 \tabularnewline
Total activities & 4 & 3 & 10 & 9 \tabularnewline
Fps & 30 & 30 & 30 & 30 \tabularnewline
Resolution & $1920\mathrm{\times}1080$ & $3840\mathrm{\times}2160$ & $640\mathrm{\times}480$ & $640\mathrm{\times}480$ \tabularnewline
\bottomrule
\end{tabular}
\end{adjustbox}
\end{table}

\subsection{Statistical Comparison of Datasets}
We perform a statistical comparison between $\newdatasetshort$ and existing popular HOI datasets, as shown in Tab.~\ref{tab:dataset_comparison}. $\newdatasetshort$ includes scenarios with three people interacting and 17 sub-activities, which is higher than any other listed dataset, standing out for its complexity and richness. With a total of 53,604 frames across 120 videos, nearly half (20,100 frames) capture concurrent partial interactions, offering a unique focus on dynamic multi-person interactions absent in other datasets. Additionally, the high video resolution ($1920\times1080$) ensures detailed feature capture, essential for advanced HOI analysis. In contrast, although Bimanual Actions is large, it is limited to dual-hand movements of an individual, leading to a more monotonic data distribution.

\section{Methodology}
We propose a bottom-up approach to design $\newmodelshort$, which (1) preserves fine-grained entity characteristics during feature fusion, and (2) progressively builds from entity-specific representations to interaction-level understanding. The bottom-up approach has been widely used in pose estimation \citep{kresovic2021bottomreview,wang2022regularizing,wang2022lite} and object detection \citep{zhou2019extremeNet,wang2019topbottom,samet2020houghnet} tasks with considerable performances. It ensures a thorough understanding of the fundamental aspects of each entity before delving into complex entity-level interactions. This approach, starting from basic features and building upwards, enables detailed feature integration to achieve more effective entity interaction analysis.

\begin{figure}
\centering
\includegraphics[width=\linewidth]{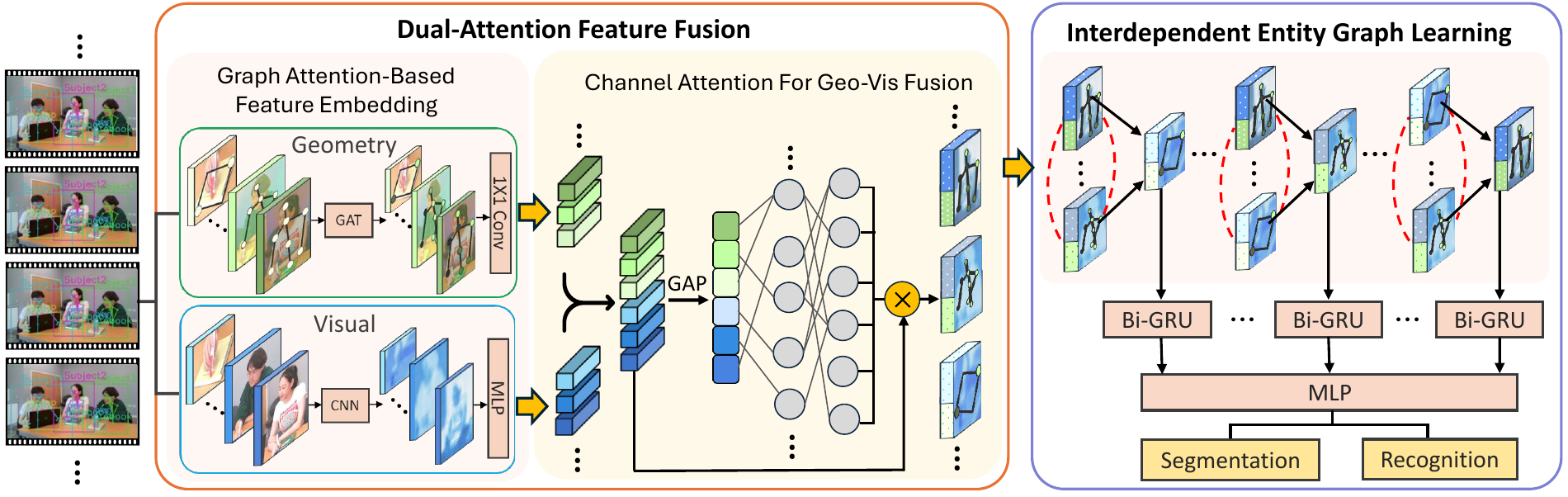}
\caption{Overview of our bottom-up framework $\newmodelshort$. We first design a dual-attention fusion for entity feature optimization, which embeds and fuses visual and geometric features in a graph attention-based mechanism and channel attention module, respectively. The enriched entity-specific representations are then inputted into the interdependent entity graph to further model explicit interactions and implicit interdependencies. Finally, we apply a BiGRU to capture the temporal dependencies to obtain segmentation and recognition results.}
\label{fig:framework}
\end{figure}

Alternative designs perform suboptimally. A top-down approach \citep{zheng2023deep}, which prioritizes a broad view of entity-level relationships before refining specific entity features, often overlooks crucial interaction details and misaligns overarching patterns with individual interaction nuances. Besides, a mixed-fusion method \citep{qiao2022geometric} that integrates feature fusion and entity interaction learning within a single graph entangles entity concepts, lacking a specific feature to represent each entity, which fails to learn HOIs explicitly. We compare these alternative architectures with our method in Experimental Results \ref{sec:experiments}.

\subsection{Dual-Attention Fusion for Feature Optimization}
Previous HOI recognition approaches primarily rely on CNN or 3D-CNN models to process visual inputs. These models extract spatio-temporal features that capture rich appearance and contextual cues from humans and objects \citep{maraghi2019zero,le2020bist,morais2021learning}. While effective in clean settings, these methods are highly sensitive to occlusions and struggle when visual cues are incomplete or ambiguous. Without explicit spatial reasoning, they often fail to capture the structural context of interactions. Incorporating geometric information is therefore critical for improving robustness and enabling accurate recognition in real-world HOI scenarios. Advanced methods such as 2G-GCN \citep{qiao2022geometric} attempt to integrate geometric features within a GCN framework to augment visual data. However, their fusion of collective geometric features with individual visual features risks hierarchical misalignment, fusion inefficiencies, and difficulties in entity distinction. CATS \citep{qiao2024category} also employs GCN to model geometric features but directly combines them with visual features, which may dilute their distinct contributions.

We propose a dual-attention fusion mechanism to optimize multimodal feature integration for entity representations (Fig.~\ref{fig:framework}). We first apply a graph attention mechanism to geometric features \citep{li2025unifiedST}, enabling the model to learn structured spatial representations by capturing the varying importance of neighboring entities. With these enriched geometric embeddings, we then employ a channel attention module to adaptively fuse geometric and visual features, selectively emphasizing informative channels while suppressing less relevant ones. This sequence ensures that spatial reasoning is established before feature fusion and allows the model to balance modality contributions more effectively. If channel attention is applied before relational modeling, it would risk fusing less informative geometric features and weaken the spatial reasoning capability. As a result, we obtain a well-contextualized entity representation that effectively blends geometric and visual cues, providing a robust foundation for subsequent entity interaction graph learning.

\subsubsection{Graph Attention-Based Feature Embedding}
Previous research \citep{qiao2022geometric,zhou2022human,qiao2024category} learns geometric features using GCNs, which typically apply the same convolution operation to all neighbors of a node. This approach fails to account for the different roles or importance that neighbors may have in the context of multi-person HOIs. This may lead to a homogenization of features that fails to capture the complexity of multi-entity dynamics.

We propose a Graph Attention Network (GAT) \citep{brody2021gatv2} based embedding to capture the evolving significance of interactions \citep{li2022multiclass_sgcn,li2025unifiedST,li2025bp-sgcn}. It learns multi-entity geometric features, adaptively weighting the importance of each entity's geometric features through an attention mechanism. 
This enables the model to expertly handle occlusions and dynamic environments for multi-person HOI recognition.

For feature representation, we concatenate the position and velocity of all entities into keypoint channels, forming geometric features $\mathcal{G} = \{g_t^{e,k}\}_{t=1,e=1,k=1}^{T,E,K} \in \mathbb{R}^4$ with $g_t^{e,k}$ as the $k$-th type features for entity $e$ at frame $t$, where $T$ denotes the total number of frames in the video, $E$ and $K$ denote the total number of entities and keypoints of an entity in a frame, respectively. Human joints and object bounding box diagonals are extracted as keypoints.

We adaptively infer spatial correlations with our GAT among keypoints $k_1$ and $k_2$ for a single timestep among entities as follows:
\begin{equation}
    \mathbf{g}^{s}_t = \alpha_{{k_1},{k_1}}\mathbf{\Theta}\mathbf{g}_{t,{k_1}} +
    \sum_{k_2 \in \mathcal{K}} \alpha_{{k_1},{k_2}}\mathbf{\Theta}\mathbf{g}_{t,{k_2}},
\end{equation}
and the attention coefficients $\alpha_{{k_1},{k_2}}$ are computed as:
\begin{equation} 
    \alpha_{{k_1},{k_2}} =
    \frac{
    \exp\left(\Gamma\left(\mathbf{a}^{\top}
    [\mathbf{\Theta}\mathbf{g}_{k_1} \, \Vert \, \mathbf{\Theta}\mathbf{g}_{k_2}]
    \right)\right)}
    {\sum_{k_3 \in \mathcal{K} \cup \{ k_3 \}}
    \exp\left(\Gamma\left(\mathbf{a}^{\top}
    [\mathbf{\Theta}\mathbf{g}_{k_1} \, \Vert \, \mathbf{\Theta}\mathbf{g}_{k_3}]
    \right)\right)},
\end{equation}
where $\mathbf{\Theta}$ and $\Gamma$ are the transformation function and LeakyReLU activation, respectively.
 
To efficiently integrate spatial and temporal information, we further process the attention-enhanced geometric features. In particular, $\mathbf{g}^{s}_t$ is then fused with a 1 × 1 convolution along the temporal channel to form spatial-temporal geometric features $\mathbf{g}^{st}_t \in \mathbb{R}^{T \times NK \times C_1}$, effectively summarizing temporal dynamics while avoiding the complexities of 3D convolutions. It is then reshaped to $\mathbf{g}^{st}_t \in \mathbb{R}^{T \times N \times KC_1}$ and embedded by a Multi-Layer Perceptron (MLP) to get entity geometric features $\mathbf{g}^{\prime}_t \in \mathbb{R}^{T \times N \times C_2}$. 

Unlike geometric features, visual features in videos contain rich contextual information and fundamental feature representations. Following \citep{morais2021learning,qiao2024category}, we extract entity visual features $\mathbf{v}_{t,n} \in \mathbb{R}^{2048}$ from ROI pooled 2D bounding boxes of humans and objects in videos, utilizing a pre-trained Faster R-CNN \citep{ren2016faster} module on the Visual Genome \citep{krishna2017visual}. They are subsequently aligned dimensionally with geometric features to $\mathbf{v}^{\prime}_t \in \mathbb{R}^{T \times N \times C_2}$ through an MLP with learnable embeddings.

\subsubsection{Geo-Vis Channel Attention-Based Feature Fusion}
Incorporating geometric and visual features poses a significant challenge due to their inherent representation and scale discrepancies. Prior approaches have attempted multimodal fusion by element-wise addition \citep{zhou2022human} or feature concatenation \citep{wan2019pose,qiao2024category}. However, such direct operations are infeasible for our task as they do not account for the disparate nature of feature spaces, leading to suboptimal learning outcomes.

We propose a novel geometry-visual channel attention-based feature fusion to effectively integrate geometric and visual features of all humans and objects, which achieves selective feature enhancement and encourages complementarity between multimodal features. We exploit channel attention mechanisms \citep{hu2018squeeze,li2024rapidseg} in geometry-visual channels of all entities. This allows the model to adaptively emphasize informative features while suppressing less relevant ones, which is especially beneficial for learning more representative visual and geometric features in diverse HOI scenarios. For instance, visual features often suffer in noisy backgrounds but thrive in scenarios with small backgrounds. Geometric features demonstrate strength in addressing partial occlusions \citep{qiao2022geometric}, which is a common situation in multi-person HOI scenarios.

Specifically, as shown in Fig.~\ref{fig:framework}, our channel attention based feature fusion module first concatenates $\mathbf{g}^{\prime}_t$ and $\mathbf{v}^{\prime}_t$ along the entity dimension to entity geometry-visual features $\mathbf{gv}_t \in \mathbb{R}^{T \times 2N \times C_2}$, and compute a channel attention $A$ as:
\begin{equation}
A = \sigma\left(\mathbf{W}_{2} \delta\left(\mathbf{W}_1 \left(GAP\left(\mathbf{gv}_t\right) \right) \right)\right),
\end{equation}
where GAP denotes Global Average Pooling \citep{lin2014network}, $\delta$ and $\sigma$ represent the ReLU and Sigmoid activation. $\mathbf{W}_1$ and $\mathbf{W}_2$ are weights of Fully-Connected (FC) layers, shared across all entities and timesteps to ensure consistent transformation and improved generalization with fewer parameters.
Apply these values to original features for attended geometry-visual fusion features:
\begin{equation}
\mathbf{gv'}_t = A \cdot \mathbf{gv}_t.
\end{equation}

Finally, we enhance the feature representation of each entity. In particular, after assigning distinct weights to each geometry-visual channel of an entity, the weighted features are strategically split into separate geometric and visual streams. These are then adeptly fused back together, producing a new enriched entity representation $\widetilde{\mathbf{gv}}_t \in \mathbb{R}^{T \times N \times C_3}$. This refined feature fusion set, being a weighted and well-contextualized blend of geometric and visual cues, sets the stage for more discerning entity interaction graph learning. 

Compared to our attention-based feature fusion, \citet{zhang2022exploring,tu2023agglomerative} apply Transformer to fuse geometric and visual features in image-based HOI detection, which is constrained in processing video data due to memory inefficiency. Graph-based feature fusion treats multimodal features as graph nodes \citep{liang2020LaneGCN, gao2020vectornet}, which is heavily reliant on the design of graph representation. As HOI is a dynamic process, it is non-trivial to manually define an appropriate representation.

\subsection{Interdependent Entity Graph} 
In HOI analysis, the majority of approaches \citep{wang2021spatio,morais2021learning,wang2023exploring} construct an independent entity graph that assumes a fixed structure to decipher spatial interactions between entities focusing solely on visual features. For example, 2G-GCN \citep{qiao2022geometric} represents geometric features of all entities as a single entity linked with visual features of object entities, failing to explicitly model interactions between all entities. CATS \citep{qiao2024category} learns interactions between human and object categories but neglects relationships between entities within the same category, which is particularly limiting in multi-person HOI scenarios.

Our insight is that an effective entity interaction graph should not only capture explicit interactions among independent entities, but also concurrently discern the implicit interdependencies that exist among neighboring entities surrounding a specific entity. These complementary focuses are crucial for understanding the intricate graph network of relations that exist around any specific entity within the scene.

To this end, we propose an interdependent entity graph to capture the interdependencies among all neighboring nodes around a particular entity with fused geometric and visual features. To improve the precision of interaction modeling and the representation of relational dynamics, we further refine it by employing attention weights between the entity in focus and its neighbors (Fig.~\ref{fig:framework} right). This entity-level graph offers a richer representation of spatial interactions in multi-person HOI scenarios, advancing the understanding of complex behavioral patterns beyond the reach of previous methods.

\begin{figure}[h!]
\centering
\includegraphics[width=0.8\linewidth]{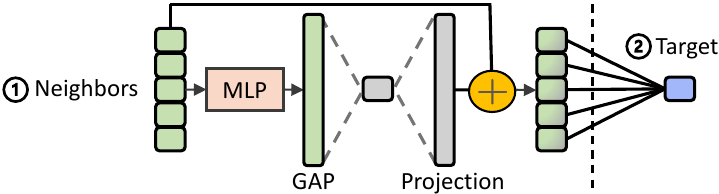}
\caption{In the interdependent entity graph, we first model neighbor features \ding{172} before aggregating them to the target entity \ding{173}.}
\label{fig:IEG}
\end{figure}

Specifically, as illustrated in Fig.~\ref{fig:IEG}, given a specific entity ${e}$ at each frame ${t}$, we first calculate the features from its neighbor $u$ to itself as follows:
\begin{equation}
     S_{t}^{u} = \lambda \times \widetilde{\mathbf{gv}}^{u}_t + (1-\lambda) \times \frac{(\text{GAP}(\mathbf{W}_3(\widetilde{\mathbf{gv}}^{u}_t)))}{N-1},
\end{equation}
where $\lambda$ controls the contextual fusion threshold and is fixed to 0.5. This value is selected based on preliminary validation experiments, and we find that the model's performance remains stable despite small variations in this setting. $\mathbf{W}_3$ is the weight of a FC layer. These neighboring features are then aggregated into a robust representation that encapsulates the collective attributes of the neighboring group:
\begin{equation}
S_{t}^{e} = STACK_{u\in N, u\neq e}(S_{t}^{u}\odot M(S_{t}^{u})),
\end{equation}
where $M(\cdot)$ is the mask indicator for valid neighbors and $\odot$ denotes element-wise multiplication. Meanwhile, we employ a dot-product attention mechanism \citep{vaswani2017transformer, morais2021learning} to obtain the attention weights between node $e$ and its neighbors as:
\begin{equation}
W_{t}^{e} =  {\textstyle \sum_{u \in N, u \neq  e}}Softmax(\frac{S_{t}^{e} (S_{t}^{u})^{T}}{\sqrt d}),
\end{equation}
where $d$ is the feature dimension. Finally, the refined feature representation of the entity is $F_{t}^{e} = W_{t}^{e} \odot S_{t}^{e}$, ensuring a contextually aware integration of features that strengthens the entity's representation within its surroundings.

To enable precise and adaptable delineation of sub-event lengths in video sequences, after obtaining the fused features of each entity at each time step, we employ a Gumbel-Softmax module \citep{jang2016gsm} to $F_{t}^{e}$. It efficiently facilitates gradient-based learning and ensures probabilistically coherent segmentation, essential for handling the dynamic nature of video data. Finally, we apply a Bi-directional Gated Recurrent Unit (BiGRU) \citep{chung2014gru} to capture the temporal dependencies between each sub-action and then use the output features to recognize sub-activities for humans and object affordances for objects, varying according to the dataset.

\section{Experimental Results}
\label{sec:experiments}
\subsection{Datasets}
We evaluate $\newmodelshort$ on multiple datasets: $\newdatasetshort$, MPHOI-72 \citep{qiao2022geometric}, CAD-120 \citep{koppula2013learning}, and Bimanual Actions \citep{dreher2020learning}, showcasing the superior results on concurrent partial HOI, two-person, single-person and two-hand HOI recognition.

The MPHOI-72 dataset is valuable for two-person HOI recognition tasks. It contains 72 videos of 8 pairs of people performing 3 distinct activities (\textit{Cheering}, \textit{Hair cutting} and \textit{Co-working}) with 13 human sub-activities (\eg, \textit{Sit}, \textit{Approach}, \textit{Pour}). Each video showcases two participants interacting with 2-4 objects from 3 unique angles. Geometric features and human sub-activity labels are frame-wise annotated.

CAD-120 is a prominent dataset for single-person HOI recognition. It contains 120 RGB-D videos, capturing 10 distinct activities executed by 4 participants, each repeated three times. In each video, a participant interacts with 1-5 objects. The dataset provides frame-wise annotations for 10 human sub-activities (\eg, \textit{opening}, \textit{cleaning}, \textit{placing}) and 12 object affordances (\eg, \textit{openable}, \textit{cleanable}, \textit{placeable}).

The Bimanual Actions dataset is a large-scale collection of 540 RGB-D videos capturing HOIs using both hands. It documents the actions of 6 subjects who engage in 9 varied bimanual tasks, with each task performed 10 times. The dataset assigns 14 unique action labels to each hand, with frame-wise annotations for each entity within the videos.

\subsection{Implementation Details}
We follow \citet{morais2021learning} and \citet{qiao2022geometric,qiao2024category} to evaluate $\newmodelshort$ on two tasks: joint segmentation and label recognition, and label recognition given known segmentation. The first task involves segmenting the timeline of each entity and classifying segment labels in a video. The second task, an extension of the first, requires labeling pre-existing segments with known ground-truth segmentation.
We utilize the $\mathrm{F}{1}@k$ metric \citep{lea2017temporal} for evaluation, applying standard thresholds of $k=10\%$, $25\%$, and $50\%$. This metric considers a predicted action segment correct if it achieves a minimum Intersection over Union (IoU) overlap of $k$ with the ground truth. It is widely adopted in temporal segmentation research \citep{lea2017temporal,farha2019ms,morais2021learning}, particularly for its ability to handle short or partial actions commonly found in HOI scenarios by requiring a certain overlap for each segment. As a result, it offers a more fine-grained evaluation of segmentation quality, capturing both the correctness of segment boundaries and the overall alignment with the ground truth.

For dataset evaluation, we use different cross-validation protocols tailored to the characteristics of each dataset to ensure subjects in the training set do not appear in the test set. For the single-person HOI datasets, CAD-120 and Bimanual Actions, we use leave-one-subject-out cross-validation, treating each individual as a separate fold. For the two-person HOI dataset MPHOI-72, we employ leave-two-subjects-out to preserve the same principle while accounting for pairs of interacting subjects. This ensures a strict separation of subjects (or subject pairs) between training and testing. For $\newdatasetshort$, our cross-validation scheme specifies three subjects not present in the training set as the test set. 

The GeoVis-GNN framework is implemented in PyTorch and trained in two stages using the AdamW optimizer. A batch size of 16 is used across all datasets. The learning rate is set to 0.0001 for both the MPHOI datasets and the CAD-120 and Bimanual Actions datasets. Training $\newdatasetshort$, MPHOI-72, CAD-120 and Bimanual Actions on four Nvidia Titan RTX GPUs take 6, 4, 8 hours and 7 days respectively, while testing the entire set takes approximately 2, 2, 6 and 20 minutes respectively.

To capture increasingly complex features while keeping computational cost reasonable, we adopt an incremental increase in dimensionality. Specifically, we set $C_1=128$, $C_2=256$, and $C_3=512$ based on empirical experimentation to balance model capacity and efficiency. $C_1$ and $C_2$ serve as mid-level embeddings for spatio–temporal transformations, while $C_3$ enables deeper representations for modeling high-level interactions.
As Bimanual Actions has a significantly more monotonic data distribution, we set $C_2 = 32$, $C_3 = 64$ and [$S^u_t=0$].

\subsection{Quantitative and Qualitative Comparison with SOTAs}
\subsubsection{Concurrent Partial HOIs}
In the $\newdatasetshort$ dataset, $\newmodelshort$ beats ASSIGN \citep{morais2021learning}, 2G-GCN \citep{qiao2022geometric} and CATS \citep{qiao2024category} by a considerable gap (Tab.~\ref{tab:mp120}). Especially under multi-person HOI conditions, ASSIGN drops below $60\%$ in $\mathrm{F}_{1}$ metrics due to occlusions affecting visual features in HOI tasks. $\newmodelshort$ shows an improvement of about $2\%$ to $4\%$ in $\mathrm{F}_{1}@\{10, 25, 50\}$ over SOTA, demonstrating its ability to effectively handle concurrent partial interactions. This highlights the strength of its dual-attention fusion strategy and interdependent entity graph in capturing essential features and modeling stable interactions, even in the presence of unexpected occlusions and complex multi-person dynamics.

\begin{table}[H]
\caption{Joined segmentation and label recognition results on $\newdatasetshort$.\label{tab:mp120}}
\centering{}
\footnotesize
\begin{tabular}{ccccc}
\toprule 
\multirow{2}{*}{Model} &  & \multicolumn{3}{c}{Sub-activity}\tabularnewline
\cmidrule{3-5} \cmidrule{4-5} \cmidrule{5-5} 
 &  & $\mathrm{F}_{1}@10$ & $\mathrm{F}_{1}@25$ & $\mathrm{F}_{1}@50$\tabularnewline
\cmidrule{1-1} \cmidrule{3-5}  
ASSIGN &  & 58.0 $\pm$ 8.5 & 53.7 $\pm$ 7.9 & 39.1 $\pm$ 7.4\tabularnewline
2G-GCN &  & 60.7 $\pm$ 6.5 & 55.3 $\pm$ 6.9 & 39.6 $\pm$ 6.5\tabularnewline
CATS &  & 62.8 $\pm$ 2.7 & 56.7 $\pm$ 4.2 & 42.8 $\pm$ 3.9\tabularnewline
\cmidrule{1-1} \cmidrule{3-5} 
$\newmodelshort$ &  & \textbf{65.1} $\pm$ 5.2 & \textbf{59.8} $\pm$ 4.7 & \textbf{46.6} $\pm$ 5.1\tabularnewline
\bottomrule
\end{tabular}
\end{table}

\begin{figure}[H]
\centering
\includegraphics[width=\linewidth]{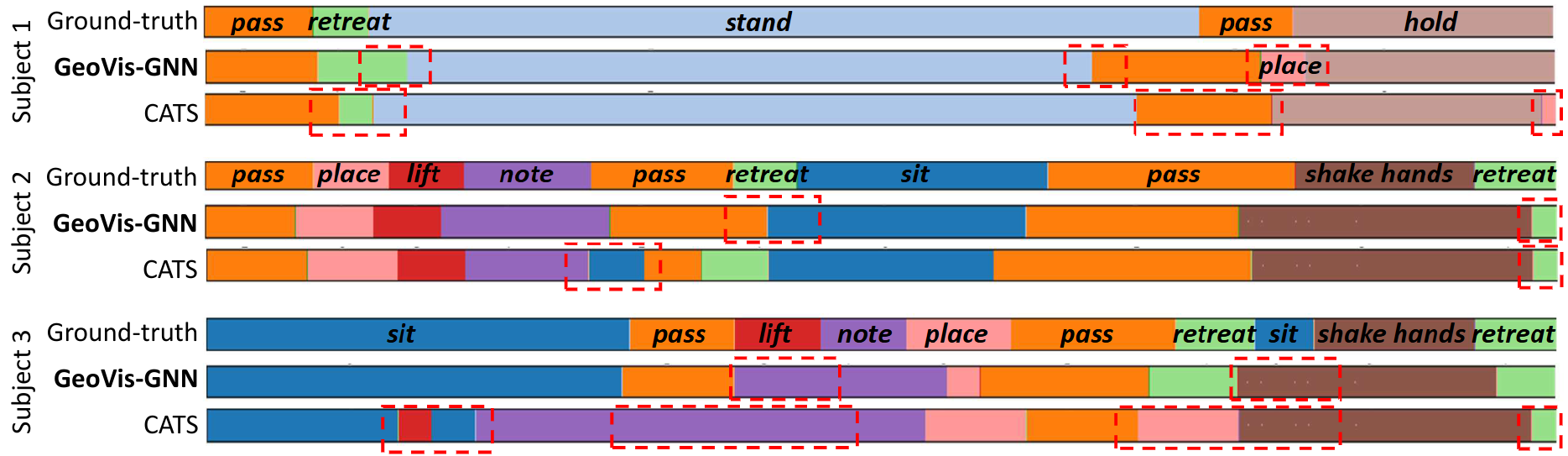}
\caption{Visualization of segmentation on $\newdatasetshort$ for \textit{Signing} activity. Red dashed boxes highlight major segmentation errors.}
\label{fig:mp120_vis}
\end{figure}

Fig.~\ref{fig:mp120_vis} illustrates the visualization results of $\newmodelshort$ and CATS on $\newdatasetshort$ comparing with Ground-truth for the \textit{Signing} activity, where red dashed boxes highlight major segmentation errors. Although both $\newmodelshort$ and CATS make errors compared to Ground-truth, $\newmodelshort$ can contribute relatively plausible segmentation results in all three subjects. For example, in subject 3, CATS oversegments \textit{sit} in the beginning and then completely misses \textit{pass} and \textit{lift} before \textit{note}, while our $\newmodelshort$ can accurately segment \textit{sit} and \textit{pass} but miss \textit{lift}. This is likely due to the \textit{lift} action of the subject being very fast and closely resembles the \textit{note} action, leading our model to misclassify \textit{lift} as \textit{note}. Incorporating temporal attention mechanisms could potentially enhance performance in the short duration of the action and its overlapping features with subsequent actions.

\subsubsection{Two-person HOIs}
$\newmodelshort$ achieves an impressive performance on the MPHOI-72 dataset (Tab.~\ref{tab:mp72}), with an $\mathrm{F}_{1}@10$ score of $84.3\%$, significantly outstripping the $71.3\%$ scored by CATS \citep{qiao2024category}. Across all $\mathrm{F}_{1}$ configurations, $\newmodelshort$ exhibits substantial improvements of $13.0\%$, $10.8\%$, and $10.6\%$, respectively. The advanced technique for fusing geometric and visual features allows to capture more complex patterns in the data, while CATS and 2G-GCN cannot leverage it due to its inefficient fusion.

\begin{table}[H]
\caption{Joined segmentation and label recognition results on MPHOI-72.\label{tab:mp72}}
\centering{}
\footnotesize
\begin{tabular}{ccccc}
\toprule 
\multirow{2}{*}{Model} &  & \multicolumn{3}{c}{Sub-activity}\tabularnewline
\cmidrule{3-5} \cmidrule{4-5} \cmidrule{5-5} 
 &  & $\mathrm{F}_{1}@10$ & $\mathrm{F}_{1}@25$ & $\mathrm{F}_{1}@50$\tabularnewline
\cmidrule{1-1} \cmidrule{3-5}  
ASSIGN &  & 59.1 $\pm$ 12.1 & 51.0 $\pm$ 16.7 & 33.2 $\pm$ 14.0 \tabularnewline
2G-GCN&  & 68.6 $\pm$ 10.4 & 60.8 $\pm$ 10.3 & 45.2 $\pm$ 6.5 \tabularnewline
CATS &  & 71.3 $\pm$ 5.0 & 65.8 $\pm$ 3.9 & 48.8 $\pm$ 5.3 \tabularnewline
\cmidrule{1-1} \cmidrule{3-5} 
$\newmodelshort$ &  & \textbf{84.3} $\pm$ 5.5 & \textbf{76.6} $\pm$ 4.5 & \textbf{59.4} $\pm$ \textbf{4.9}\tabularnewline
\bottomrule
\end{tabular}
\end{table}

\begin{figure}[H]
\centering
\includegraphics[width=\linewidth]{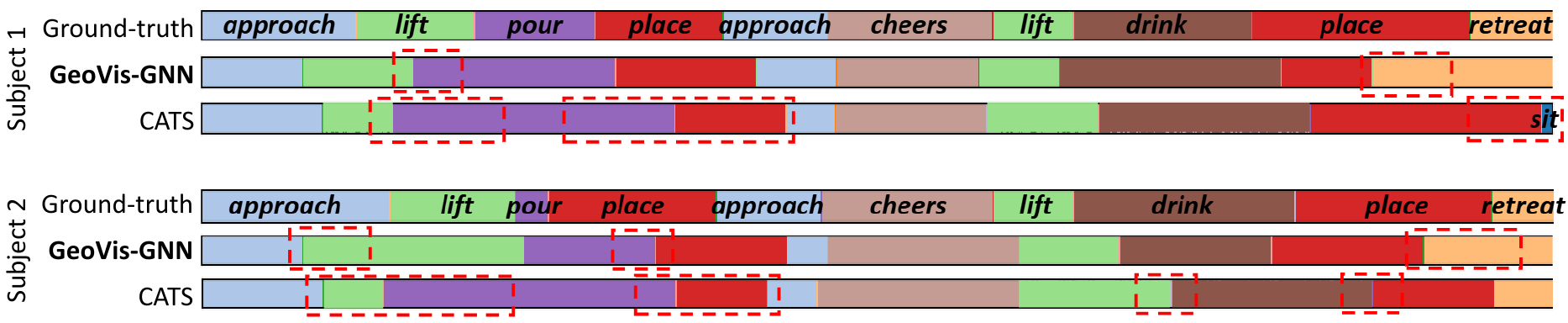}
\caption{Visualization of segmentation on MPHOI-72 for \textit{Cheering} activity. Red dashed boxes highlight major segmentation errors.}
\label{fig:mphoi72_vis}
\end{figure}

Fig.~\ref{fig:mphoi72_vis} shows the visualization of segmentation and labeling on the MPHOI-72 dataset with the two advanced models for the \textit{Cheering} activity comparing with Ground-truth. $\newmodelshort$ presents more reasonable and robust segmentation results in all sub-activities, while CATS provides some unexpected abnormal results in certain sub-activities, such as \textit{pour} and \textit{place}. Interestingly, CATS directly recognizes the static action \textit{sit} rather than the ongoing action \textit{retreat} following \textit{place} at the end of the activity for subject 1. This may result from the dominant role of visual features, as these two actions appear similar in the front view.

\subsubsection{Single-person HOIs}

\begin{table}[H]
\caption{Joined segmentation and label recognition results on CAD-120.}\label{tab:cad120}
\footnotesize
\centering{}
\begin{adjustbox}{width=1\linewidth}
\begin{tabular}{ccccccccc}
\toprule 
\multirow{2}{*}{Model} &  & \multicolumn{3}{c}{Sub-activity} &  & \multicolumn{3}{c}{Object Affordance}\tabularnewline
 &  & $\mathrm{F}_{1}@10$ & $\mathrm{F}_{1}@25$ & $\mathrm{F}_{1}@50$ &  & $\mathrm{F}_{1}@10$ & $\mathrm{F}_{1}@25$ & $\mathrm{F}_{1}@50$\tabularnewline
\cmidrule{1-1} \cmidrule{3-5} \cmidrule{7-9}
rCRF &  & 65.6 $\pm$ 3.2 & 61.5 $\pm$ 4.1 & 47.1 $\pm$ 4.3 &  & 72.1 $\pm$ 2.5 & 69.1 $\pm$ 3.3 & 57.0 $\pm$ 3.5\tabularnewline
Independent BiRNN &  & 70.2 $\pm$ 5.5 & 64.1 $\pm$ 5.3 & 48.9 $\pm$ 6.8 &  & 84.6 $\pm$ 2.1 & 81.5 $\pm$ 2.7 & 71.4 $\pm$ 4.9\tabularnewline
ATCRF &  & 72.0 $\pm$ 2.8 & 68.9 $\pm$ 3.6 & 53.5 $\pm$ 4.3 &  & 79.9 $\pm$ 3.1 & 77.0 $\pm$ 4.1 & 63.3 $\pm$ 4.9\tabularnewline
Relational BiRNN &  & 79.2 $\pm$ 2.5 & 75.2 $\pm$ 3.5 & 62.5 $\pm$ 5.5 &  & 82.3 $\pm$ 2.3 & 78.5 $\pm$ 2.7 & 68.9 $\pm$ 4.9\tabularnewline
ASSIGN &  & 88.0 $\pm$ 1.8 & 84.8 $\pm$ 3.0 & 73.8 $\pm$ 5.8 &  & 92.0 $\pm$ 1.1 & 90.2 $\pm$ 1.8 & 82.4 $\pm$ 3.5\tabularnewline
2G-GCN &  & 89.5 $\pm$ 1.6  & 87.1 $\pm$ 1.8 & 76.2 $\pm$ 2.8 &  & 92.4 $\pm$ 1.7 & \textbf{90.4} $\pm$ 2.3 & 82.7 $\pm$ 2.9\tabularnewline
CATS &  & 89.6 $\pm$ 2.1 & 87.3 $\pm$ 1.5 & 76.0 $\pm$ 3.5 &  & 90.2 $\pm$ 1.5 & 89.1 $\pm$ 2.4 & 80.5 $\pm$ 2.8\tabularnewline
\cmidrule{1-1} \cmidrule{3-5} \cmidrule{7-9}
$\newmodelshort$ &  & \textbf{89.9} $\pm$ 2.0 & \textbf{87.8} $\pm$ 1.9 & \textbf{76.7} $\pm$ 3.1 &  & \textbf{92.7} $\pm$ \textbf{0.4} & \textbf{90.4} $\pm$ \textbf{0.6} & \textbf{83.3} $\pm$ \textbf{1.8}\tabularnewline
\bottomrule
\end{tabular}
\end{adjustbox}
\end{table}

Tab.~\ref{tab:cad120} shows the effectiveness of $\newmodelshort$ in CAD-120 evaluated by sub-activity and object affordance labels. $\newmodelshort$ beats previous visual-based \citep{sener2015rcrf,koppula2016anticipating,morais2021learning} and geometry-informed \citep{qiao2022geometric,qiao2024category} networks for both labels and achieves the highest $\mathrm{F}_{1}$ scores of mean in every configuration. Notably, the two geometry-informed networks show comparable performance in human sub-activity recognition, but CATS performs poorly in object affordance recognition. This may be due to two main factors: an imbalance in feature representation, with fewer keypoints for objects than humans, reducing object emphasis in the scene graph, while the dual-attention feature fusion in $\newmodelshort$ helps mitigate this. Additionally, our task requires both segmentation and label recognition, a two-stage process that does not align well with the end-to-end framework of CATS, which may struggle with such distinct processing stages. Although CATS performs well in multi-person HOI scenarios, empirical results indicate that it is less suited for single-person HOI tasks. Therefore, in the subsequent HOI recognition comparisons involving a single individual, we use 2G-GCN as the state-of-the-art benchmark.

\begin{figure}[H]
\centering
\includegraphics[width=\linewidth]{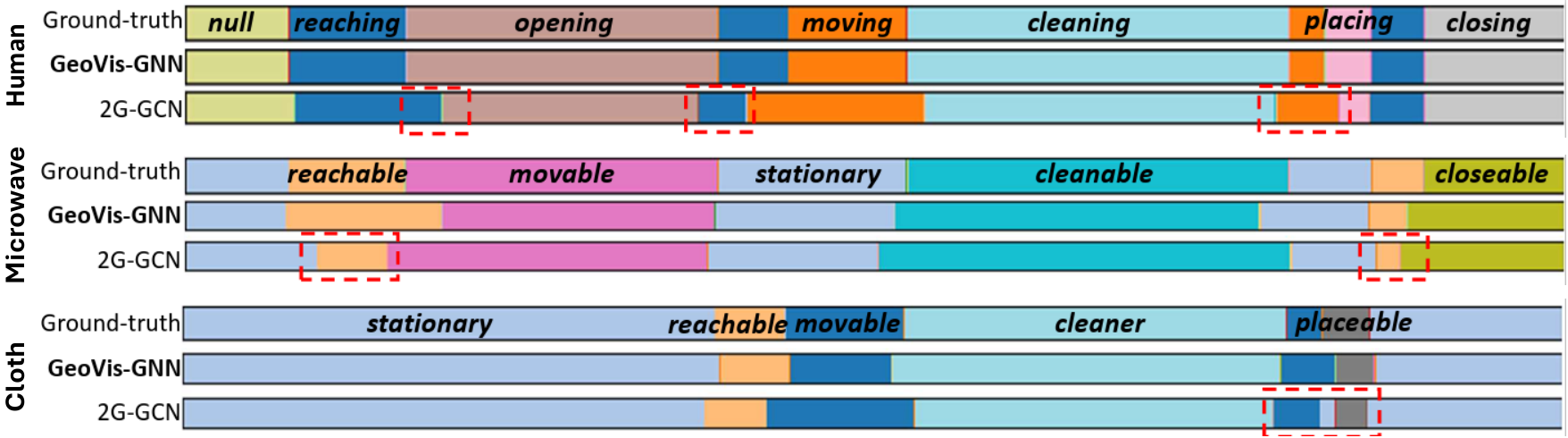}
\caption{Visualization of segmentation on CAD-120 for \textit{Cleaning objects} activity. Red dashed boxes highlight major segmentation errors.}
\label{fig:cad_vis}
\end{figure}

Fig.~\ref{fig:cad_vis} presents the visualization outcomes for the \textit{Cleaning Objects} activity in CAD-120, depicting a scene where a person uses a cloth to clean a microwave. The qualitative analysis shows that $\newmodelshort$ surpasses 2G-GCN in recognizing human sub-activities and object affordances, notably \textit{reachable} and \textit{movable} for the microwave, closely matching the Ground-truth.

\subsubsection{Two-hand HOIs} 

\begin{table}[H]
\caption{Joined segmentation and label recognition results on Bimanual Actions.\label{tab:bimacs}}
\centering{}
\footnotesize
\begin{tabular}{ccccc}
\toprule 
\multirow{2}{*}{Model} &  & \multicolumn{3}{c}{Sub-activity}\tabularnewline
\cmidrule{3-5} 
 &  & $\mathrm{F}_{1}@10$ & $\mathrm{F}_{1}@25$ & $\mathrm{F}_{1}@50$\tabularnewline
\cmidrule{1-1} \cmidrule{3-5}  
\citet{dreher2020learning} &  & 40.6 $\pm$ 7.2 & 34.8 $\pm$ 7.1 & 22.2 $\pm$ 5.7\tabularnewline
Independent BiRNN &  & 74.8 $\pm$ 7.0 & 72.0 $\pm$ 7.0 & 61.8 $\pm$ 7.3\tabularnewline
Relational BiRNN &  & 77.7 $\pm$ 3.9 & 75.0 $\pm$ 4.2 & 64.8 $\pm$ 5.3\tabularnewline
ASSIGN &  & 84.0 $\pm$ 2.0 & 81.2 $\pm$ 2.0 & 68.5 $\pm$ 3.3\tabularnewline
2G-GCN &  & 85.0 $\pm$ 2.2 & 82.0 $\pm$ 2.6 & 69.2 $\pm$ 3.1\tabularnewline
\cmidrule{1-1} \cmidrule{3-5} 
$\newmodelshort$ &  & \textbf{85.8} $\pm$ \textbf{2.2} & \textbf{82.7} $\pm$ 2.8 & \textbf{69.7} $\pm$ \textbf{3.0}\tabularnewline
\bottomrule
\end{tabular}
\end{table}

\begin{figure}[H]
\centering
\includegraphics[width=\linewidth]{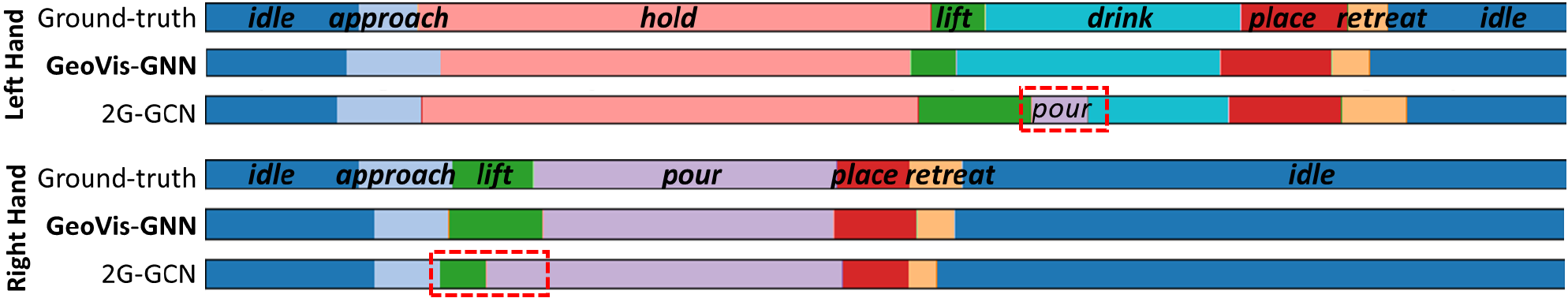}
\caption{Visualization of segmentation on Bimanual Actions for \textit{Pouring} activity. Red dashed boxes highlight major segmentation errors.}
\label{fig:bim_vis}
\end{figure}

$\newmodelshort$ achieves the superior performance on the large-scale Bimanual Actions dataset (Tab.~\ref{tab:bimacs}), with near 1\% improvement in the same standard deviation at $\mathrm{F}_{1}@10$. The slight improvement is partly due to the limited hand pose estimation that OpenPose \citep{cao2018openpose} uses for the hand skeleton of the dataset, which may introduce noise, especially in occlusions.
Fig.~\ref{fig:bim_vis} presents the visualization outcomes for the \textit{Pouring} activity in Bimanual Actions. The qualitative analysis demonstrates that $\newmodelshort$ has outstanding performance in segmenting and recognizing actions of both hands, which almost overlaps the Ground-truth, while 2G-GCN oversegments some sub-activities like \textit{pour}.

\subsection{Scenario-Based Performance and Error Analysis}

\begin{table}[ht]
\centering
\footnotesize
\caption{GeoVis-GNN performance across different HOI scenarios. The rightmost column indicates the performance gap relative to MPHOI-120.}
\label{tab:scenario_breakdown}
\begin{tabular}{lcc}
\toprule
\textbf{Dataset} & \textbf{Scenario} & \textbf{ Difference ($\mathrm{F}_{1}@10$)} \\
\midrule
CAD-120 & Single-person (General) & +24.8\%  \\
Bimanual Actions & Single-person (Bimanual)& +20.7\% \\
MPHOI-72 & Two-person (Full) & +19.2\%  \\
MPHOI-120 & Multi-person (Partial) & \multicolumn{1}{c}{--}  \\
\bottomrule
\end{tabular}
\end{table}

\begin{figure}[H]
\centering
\includegraphics[width=0.55\linewidth]{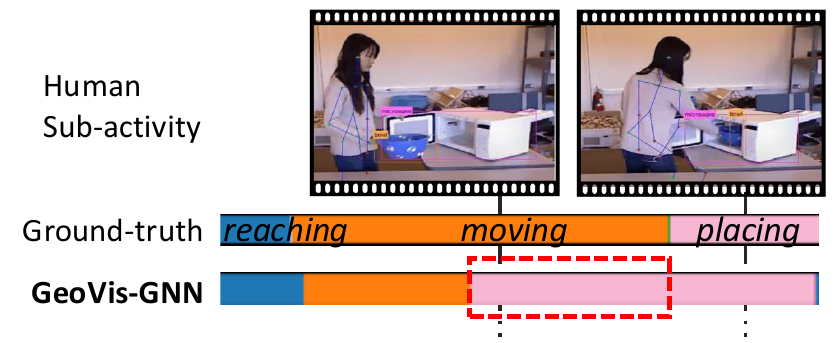}
\caption{Example of sub-activity segmentation error on the CAD-120 dataset for \textit{Microwaving Food} activity. Corresponding RGB frames are provided for visual context.}
\label{fig:cad120_microfoods}
\end{figure}

Tab.~\ref{tab:scenario_breakdown} summarizes GeoVis-GNN’s performance across different HOI scenarios with performance gaps of MPHOI-120 at $\mathrm{F}_{1}@10$. It arranges the datasets from single-person to multi-person partial, revealing a progressive increase in complexity. In single-person scenarios, CAD-120 shows the largest gap relative to MPHOI-120 at +24.8\%, indicating that single-participant tasks with clear sub-activity boundaries are relatively straightforward. Similarly, Bimanual Actions follows with a +20.7\% gap, reflecting simpler interactions than multi-person scenarios. GeoVis-GNN generally distinguishes between left- and right-hand movements effectively. Furthermore, Fig.~\ref{fig:cad120_microfoods} shows an example of sub-activity segmentation by GeoVis-GNN compared to the ground-truth, with corresponding RGB screenshots for visual reference. The segmentation error in the red box occurs because the model misclassifies part of the \textit{moving} phase as \textit{placing}, likely due to the smooth transition and similar motion patterns between the two sub-activities. This suggests that the model lacks sensitivity to subtle temporal boundaries. Improving temporal modeling or introducing boundary-aware supervision could help address this issue.

\begin{figure}[H]
\centering
\includegraphics[width=0.55\linewidth]{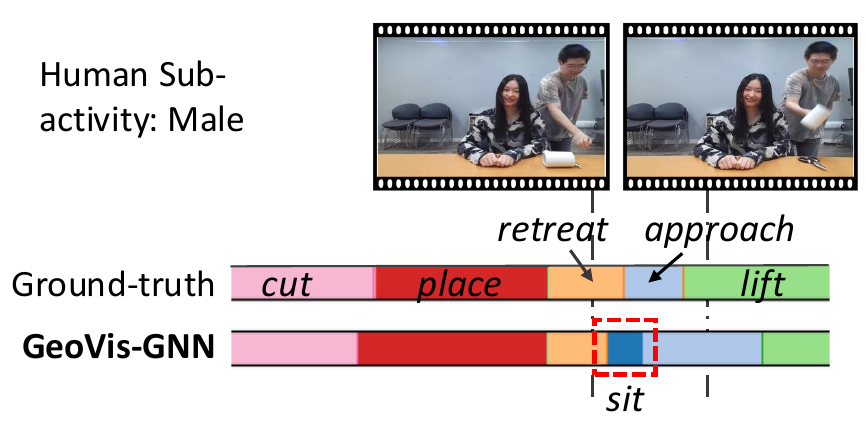}
\caption{Example of sub-activity segmentation error on the MPHOI-72 dataset for \textit{Hair Cutting} activity. Corresponding RGB frames are provided for visual context.}
\label{fig:mp72_hair}
\end{figure}

In multi-person settings, the challenges are more pronounced. MPHOI-72 focuses on two fully engaged participants and has a +19.2\% advantage. Although moderate occlusions and overlapping actions are present, the model generally maintains good performance. In contrast, MPHOI-120, which features partial engagements, idle participants, and concurrent interactions, yields a significantly lower score of 65.1\% due to heavy occlusions and ambiguous sub-activity boundaries. These results highlight the difficulty of accurately segmenting short actions and managing overlapping activities in crowded, dynamic scenes. Fig.~\ref{fig:mp72_hair} shows an over-segmentation error by GeoVis-GNN on the MPHOI-72 dataset during the \textit{Hair Cutting} activity. As highlighted in the red box, the model incorrectly inserts a \textit{sit} action between \textit{place} and \textit{approach}. This likely results from short-term pose ambiguity, causing the model to misinterpret a brief motion pause as a distinct sub-activity. This suggests the need for improved temporal smoothing to reduce false segment boundaries.

\subsection{Cross-Dataset Zero-Shot Study}
In real-world applications, models usually perform reliably on unseen data distributions without the luxury of extensive retraining or domain-specific adaptations. To demonstrate the robustness and generalization capabilities of our proposed $\newmodelshort$, we conduct a cross-dataset zero-shot evaluation, as detailed in Table~\ref{tab:zeroshot}. This study involves training $\newmodelshort$ exclusively on the concurrent partial interaction dataset and subsequently testing it on the two-person HOI dataset. 

Our results show that $\newmodelshort$ significantly outperforms the existing baselines, ASSIGN, 2G-GCN and CATS, achieving an improvement of 3.6\% in the $\mathrm{F}_{1}@10$ score. This substantial performance gain underscores the stronger generalization ability of $\newmodelshort$ compared to state-of-the-art methods. The ability to effectively transfer learned features from a more complex concurrent partial HOI scenario to a simpler two-person setting highlights the model's adaptability and transferability across diverse multi-person HOI datasets.

\begin{table}[h]
\caption{Zero-shot results of training on concurrent partial interaction dataset ($\newdatasetshort$) and testing on two-person HOI dataset (MPHOI-72).\label{tab:zeroshot}}
\centering{}
\footnotesize
\begin{tabular}{ccccc}
\toprule 
\multirow{2}{*}{Model} &  & \multicolumn{3}{c}{Sub-activity}\tabularnewline
\cmidrule{3-5}
 &  & $\mathrm{F}_{1}@10$ & $\mathrm{F}_{1}@25$ & $\mathrm{F}_{1}@50$\tabularnewline
\cmidrule{1-1} \cmidrule{3-5}  
ASSIGN &  & 33.7 & 31.5 & 28.2 \tabularnewline
2G-GCN &  & 36.2 & 33.3 & 30.4 \tabularnewline
CATS &  & 38.5 & 35.6 & 33.2 \tabularnewline
\cmidrule{1-1} \cmidrule{3-5} 
\textbf{$\newmodelshort$} &  & 42.1 & 40.3 & 34.5\tabularnewline
\bottomrule
\end{tabular}
\end{table}

Additionally, in many real-world scenarios, target domain fine-tuning or transfer learning is often employed to adapt models to specific environments. However, our zero-shot results, while not reaching the performance levels achievable when training and testing on the same two-person dataset, are achieved without any such fine-tuning, relying solely on training with one dataset and testing on another that do not necessarily share a direct relationship. This suggests that $\newmodelshort$ has the potential to generalize across different datasets with varying characteristics, even without extensive retraining. Although there is room for improvement, the results are promising and indicate that our approach can still be valuable in scenarios where labeled data for every possible situation may not be readily available.

\subsection{Ablation Study and Alternative Architecture}

\begin{figure}[H]
\centering
\includegraphics[width=0.9\linewidth]{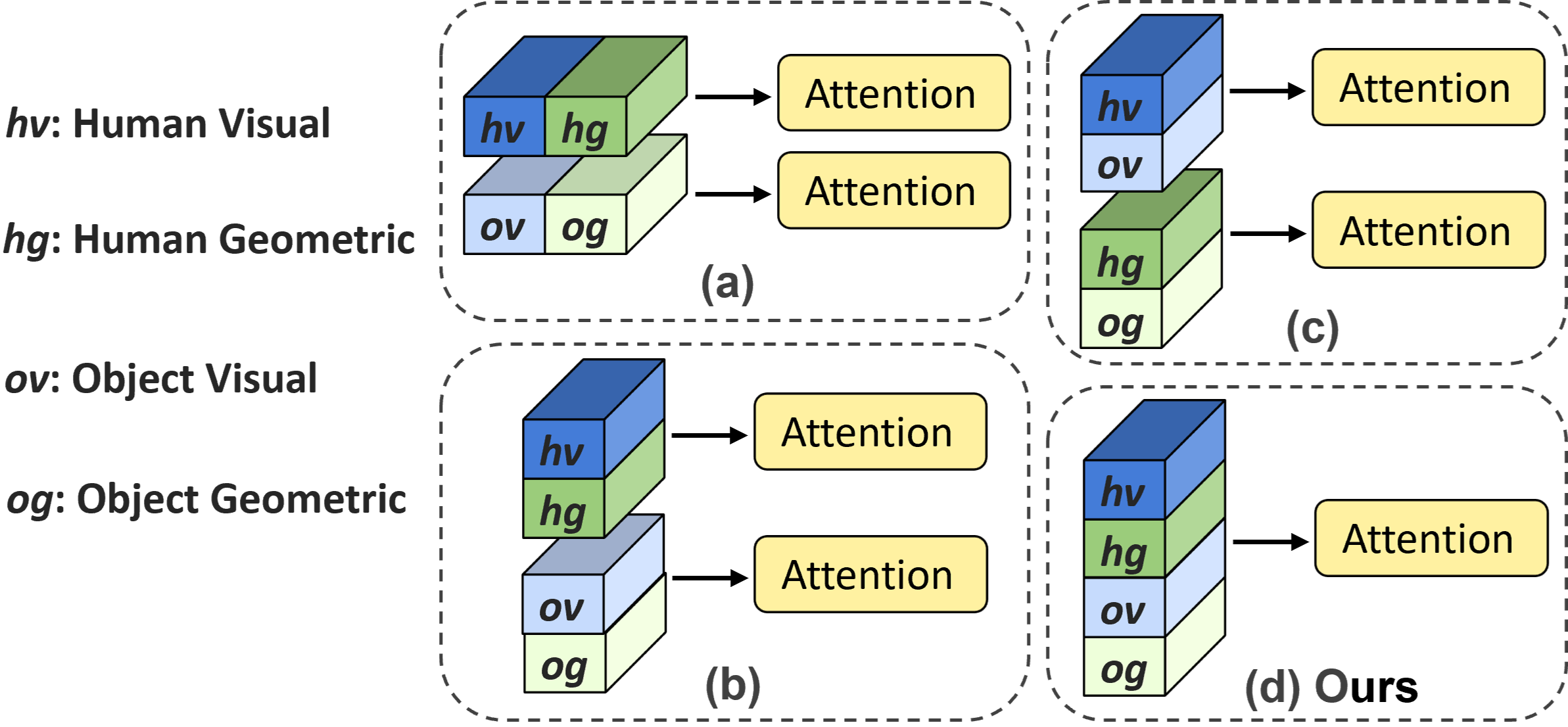}
\caption{Different designs to combine geometric and visual features in channel attention-based feature fusion.}
\label{fig:fusion_design}
\end{figure}

We extensively evaluate the design of channel attention-based feature fusion. Fig.~\ref{fig:fusion_design} shows four design strategies, in which: \textbf{(a)}: Separately concatenate human features \textit{hv, hg} and object features \textit{ov, og} on feature-channel with attentions; \textbf{(b)}: Separately concatenate human features \textit{hv, hg} and object features \textit{ov, og} on entity-channel with attentions; \textbf{(c)}: Separately concatenate visual features \textit{hv, ov} and geometric features \textit{hg, og} on entity-channel with attentions; \textbf{(d) Ours}: Concatenate all features \textit{hv, hg, ov, og} on entity-channel with a unified attention. The results of the comparison are shown in Tab.~\ref{tab:fusion_design}. Our design \textbf{(d)} presents the highest $\mathrm{F}_{1}$ score with a significant improvement gap \textit{w.r.t.} other designs. Notably, design \textbf{(a)} shows the lowest score, indicating the importance of entity-channel fusion. Although \textbf{(b)} and \textbf{(c)} contribute relatively high score, they still show 3.7\% and 6\% performance degradation in $\mathrm{F}_{1}@10$, respectively. This demonstrates the efficiency of our holistic entity-channel attention in selectively enhancing the most crucial visual or geometric features among all entities.

\begin{table}[H]
\caption{Results of different strategies in  channel attention-based feature fusion on $\newdatasetshort$.\label{tab:fusion_design}}
\footnotesize
\centering{}
\begin{adjustbox}{width=0.8\linewidth}
\begin{tabular}{lccc}
\toprule 
\multirow{2}{*}{Model} & \multicolumn{3}{c}{Sub-activity}\tabularnewline
\cmidrule{2-4} 
  & $\mathrm{F}_{1}@10$ & $\mathrm{F}_{1}@25$ & $\mathrm{F}_{1}@50$\tabularnewline
 \cmidrule{1-4} 
(a) ho feature-channel attention & 58.2 $\pm$ 4.0 & 50.7 $\pm$ 4.2 & 38.4 $\pm$ 3.6\tabularnewline
(b) ho entity-channel attention & 61.4 $\pm$ 5.7 & 56.5 $\pm$ 5.3 & 40.4 $\pm$ 4.7\tabularnewline
(c) vg entity-channel attention & 59.1 $\pm$ 5.3 & 50.4 $\pm$ 6.0 & 39.9 $\pm$ 4.8\tabularnewline
 \cmidrule{1-4} 
(d) \textbf{$\newmodelshort$} (ours) & 65.1 $\pm$ 5.2 & 59.8 $\pm$ 4.7 & 46.6 $\pm$ 5.1\tabularnewline
\bottomrule
\end{tabular}%
\end{adjustbox}
\end{table}

\begin{table}[H]
\caption{Architecture alternative and ablation study on $\newdatasetshort$. CAF and IEG denote the channel attention-based fusion and the interdependent entity graph, respectively.\label{tab:ablation}}
\centering{}\resizebox{0.8\textwidth}{!}{%
\begin{tabular}{lccc}
\toprule 
\multirow{2}{*}{Model} & \multicolumn{3}{c}{Sub-activity}\tabularnewline
\cmidrule{2-4} 
 & $\mathrm{F}_{1}@10$ & $\mathrm{F}_{1}@25$ & $\mathrm{F}_{1}@50$\tabularnewline
\cmidrule{1-4} 
(1) GAT, w CAF, w/o IEG & 61.2 $\pm$ 6.0 & 55.7 $\pm$ 5.2 & 45.4 $\pm$ 4.6\tabularnewline
(2) GAT, w/o CAF\&IEG & 59.3 $\pm$ 6.1 & 52.5 $\pm$ 5.7 & 39.4 $\pm$ 4.3\tabularnewline
(3) GCN, w/o CAF\&IEG & 58.6 $\pm$ 6.4 & 51.5 $\pm$ 5.3 & 38.3 $\pm$ 5.7\tabularnewline
(4) Top-down architecture & 62.8 $\pm$ 5.7 & 56.7 $\pm$ 5.2 & 42.8 $\pm$ 4.9\tabularnewline
\cmidrule{1-4} 
(5) \textbf{$\newmodelshort$} (ours) & 65.1 $\pm$ 5.2 & 59.8 $\pm$ 4.7 & 46.6 $\pm$ 5.1\tabularnewline
\bottomrule
\end{tabular}}
\end{table}

To further validate the effectiveness and complementary roles of each module, we conduct ablation studies on $\newdatasetshort$, where CAF and IEG refer to the channel attention-based fusion and the interdependent entity graph, respectively (Tab.~\ref{tab:ablation}). Specifically, variant (1) removes IEG, variant (2) removes both CAF and IEG, variant (3) removes CAF and IEG while replacing the GAT-based geometric embedding with GCN, and variant (4) adopts an alternative top-down design instead of our bottom-up architecture.

Our results show that removing any module leads to a significant performance drop, confirming that each component not only addresses a specific challenge but also enhances the entire pipeline. For instance, variant (1) sees a 3.9\% decline in $\mathrm{F}_{1}@10$, highlighting the critical role of IEG in modeling complex entity interactions. Likewise, variants (2) and (3) drop by 5.8\% and 6.5\%, respectively, underscoring the importance of CAF for effectively merging geometric and visual features, and the GAT for generating expressive geometric embeddings. Moreover, comparing the top-down approach (variant (4)) to our final method (5) reveals that the bottom-up framework better integrates multimodal information and preserves fine-grained entity details. These findings collectively demonstrate the synergy among GAT, CAF, and IEG, where each module contributes to robust HOI recognition by providing refined features that the subsequent modules further leverage, resulting in more accurate segmentation and interaction understanding.

\subsection{HOI Attention Analysis}

\begin{figure}[H]
\centering
\includegraphics[width=\linewidth]{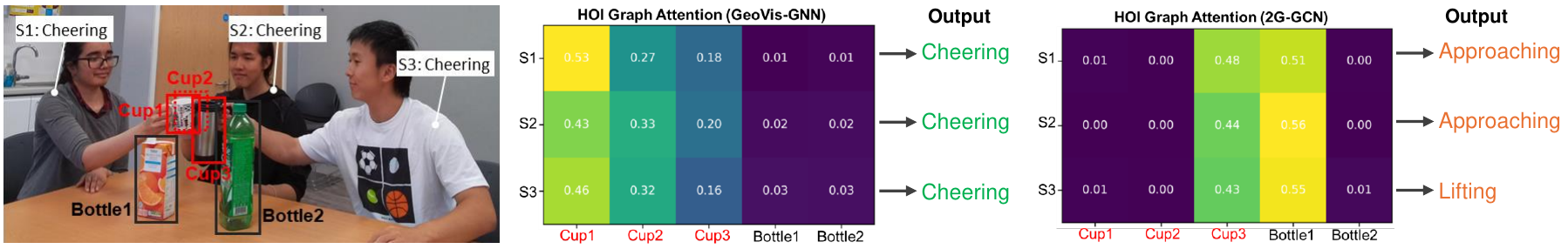}
\caption{Visualization of HOI attention maps for GeoVis-GNN and 2G-GCN during a “Cheering" activity. Correct and incorrect recognition results are highlighted in green and orange, respectively.}
\label{fig:attmap}
\end{figure}

To enhance the interpretability of our model, we deep into the attention analysis in the HOI graph. We compare GeoVis-GNN with the recent advanced method that constructs entity-level HOI graphs. Fig.~\ref{fig:attmap} presents a comparative analysis of HOI attention maps in entity-level graphs generated by GeoVis-GNN and 2G-GCN for a \textit{Cheering} activity involving three subjects, each holding a cup, with two bottles placed on the table. In the left attention map, our GeoVis-GNN model demonstrates its superior interpretability by accurately focusing on all three cups, even effectively handling occlusions, such as Cup2 being partially hidden behind Cup1. This targeted attention enables the model to correctly recognize the \textit{Cheering} sub-activity for all three subjects (highlighted in green).

In contrast, the 2G-GCN model exhibits less precise attention, incorrectly focusing on Cup3 and Bottle1, leading to erroneous sub-activity predictions such as \textit{Approaching} and \textit{Lifting} (highlighted in orange). This comparison highlights GeoVis-GNN's ability to maintain robust attention across relevant entities, even in occluded or cluttered environments, thereby ensuring more accurate HOI recognition. The clear distinction in attention focus between the two models underscores the effectiveness of our bottom-up approach in capturing the essential elements of complex interactions, which is critical for accurate activity recognition in multi-person scenarios.

\subsection{Analysis of Varying Number of Objects}
Tab.~\ref{tab:2ob_mp120} presents a comprehensive analysis of our model's performance when varying the number of objects considered on the $\newdatasetshort$ dataset. Notably, $\newdatasetshort$ contains 2-5 objects in total, even when using only 2 objects, our model outperforms the 2G-GCN baseline, demonstrating its robustness and highlighting its capability to extract meaningful interactions even from a limited set of objects.

Increasing the number of objects from 2 to 5 improves performance across all $\mathrm{F}_{1}$ metrics, but also increases memory cost. This trade-off suggests that while more objects provide richer interaction contexts, leading to better recognition accuracy, the memory requirements scale with the number of objects included. However, in highly cluttered environments with potentially hundreds of objects, our design offers an advantage by enabling the selection of a fixed number of objects to avoid a linear increase in memory consumption.

\begin{table}[H]
\caption{Results of different number of object usage on $\newdatasetshort$.\label{tab:2ob_mp120}}
\centering{}
\footnotesize
\begin{tabular}{ccccc}
\toprule 
\multirow{2}{*}{Model} &  & \multicolumn{3}{c}{Sub-activity}\tabularnewline
\cmidrule{3-5} \cmidrule{4-5} \cmidrule{5-5} 
 &  & $\mathrm{F}_{1}@10$ & $\mathrm{F}_{1}@25$ & $\mathrm{F}_{1}@50$\tabularnewline
\cmidrule{1-1} \cmidrule{3-5}
2 objects only &  & 61.4 $\pm$ 3.4 & 55.4 $\pm$ 2.0  & 40.1 $\pm$ 3.2\tabularnewline
3 objects only &  & 62.6 $\pm$ 6.9 & 56.2 $\pm$ 8.2  & 41.8 $\pm$ 9.1\tabularnewline
4 objects only &  & 63.1 $\pm$ 6.4 & 56.7 $\pm$ 7.5  & 43.2 $\pm$ 8.7\tabularnewline
\cmidrule{1-1} \cmidrule{3-5} 
\textbf{$\newmodelshort$} (5 objects) &  & 65.1 $\pm$ 5.2 & 59.8 $\pm$ 4.7 & 46.6 $\pm$ 5.1\tabularnewline
\bottomrule
\end{tabular}
\end{table}

\subsection{Parameter Size and Inference Time Analysis}

\begin{table}[H]
\caption{Comparison of parameter size (M), inference time (millisecond per frame) and performance between $\newmodelshort$ and state-of-the-arts on $\newdatasetshort$.}
\label{tab:param_size_time_analysis}
\centering
\renewcommand{\arraystretch}{1.1}
\footnotesize
\begin{tabular}{@{}cccccc@{}}
\toprule 
\multirow{2.5}{*}{Model} &  & Param & Time & & \multicolumn{1}{c}{Sub-activity}\tabularnewline
\cmidrule{6-6} 
 &  & \footnotesize{(M)} & \footnotesize{(ms/f)} & & $\mathrm{F}_{1}@10$ \tabularnewline
\cmidrule{1-1} \cmidrule{3-4} \cmidrule{6-6}
ASSIGN&  & 121 & 79 & & 58.0 $\pm$ 8.5 \tabularnewline
2G-GCN&  & 148 & 84 & & 60.7 $\pm$ 6.5 \tabularnewline
CATS &  & 132 & 137 & & 62.8 $\pm$ 2.7 \tabularnewline
\cmidrule{1-1} \cmidrule{3-4} \cmidrule{6-6}  
$\newmodelshort$  & & 130 & 119 & & \textbf{65.1} $\pm$ 5.2 \tabularnewline
\bottomrule
\end{tabular}
\end{table}

To verify the efficiency of our approach, Tab.~\ref{tab:param_size_time_analysis} compares GeoVis-GNN against 2G-GCN, CATS, and ASSIGN, all of which represent recent state-of-the-art HOI recognition frameworks, on the $\newdatasetshort$ dataset. Specifically, 2G-GCN and CATS combine geometric and visual cues, while ASSIGN only focuses on visual features. GeoVis-GNN demonstrates greater efficiency with a smaller parameter size (130M) compared to 2G-GCN (148M) and CATS (132M), while achieving competitive interactive times of 119 ms/f. Additionally, GeoVis-GNN achieves a notable performance improvement, underscoring its capability to balance efficiency and accuracy effectively in multi-person HOI recognition tasks.

\section{Conclusion and Discussion}
Our bottom-up $\newmodelshort$ framework for video-based multi-person HOI recognition introduces a novel dual-attention fusion mechanism. It optimizes feature integration by embedding and fusing visual and geometric features using a graph attention mechanism followed by a channel attention module. These enhanced entity-specific representations are then fed into an interdependent entity graph, enabling the modeling of both explicit interactions and implicit interdependencies for a more comprehensive understanding of multi-person HOI. Additionally, we propose a challenging concurrent partial interaction dataset and $\newmodelshort$ sets new benchmarks across various HOI scenarios.

Our attention-based feature fusion effectively handles scenes with multiple entities by discerning dynamic relevance and underlying connections among individuals. In highly cluttered environments — where dozens of people or objects may overlap — the root issue is that key interactions, whether contact-based or not, risk being overwhelmed by irrelevant visual clutter. For instance, a person watching TV in a room filled with other objects and people may go unnoticed if the system cannot separate important cues from background noise. This interplay between partial interactions and large-scale clutter underscores a deeper need for efficient extraction of both in-contact and non-contact interactions \citep{jiang2013hallucinated,hassan2021populating,nie2022pose2room}. Identifying the most probable HOIs in such scenarios requires robust methods for filtering out extraneous information and focusing on contextually meaningful entities.

While our concurrent partial interaction dataset closely reflects real-world multi-person HOIs, its controlled indoor settings do not fully mirror the unpredictability of in-the-wild situations. The underlying cause is that real-world environments often introduce variables like inconsistent lighting, unpredictable occlusions, diverse camera angles, and partially missing objects \citep{tripathi2023deco,yang2024lemon,ye2023human}. These factors, compounded by more fluid participant behaviors, lead to greater data ambiguity and annotation difficulty. Although we capture significant variation in our dataset, future work will extend to in-the-wild HOI videos. Tackling these unstructured real-world contexts requires innovative strategies to handle sudden motion, incomplete viewpoints, and other complexities beyond the scope of indoor, well-annotated data.

Contemporary HOI recognition often depends on precise, frame-level annotations \citep{li2021weakly}, which become costly and inconsistent when interactions are frequent and subtle — common traits in multi-person environments. The core problem is that a large volume of overlapping sub-activities escalates labeling complexity, amplifying human errors and making the labeling process time-consuming. Moreover, ambiguous transitions (e.g., partial engagement or fleeting interactions) make it hard for annotators to decide when a sub-activity starts or ends. Weakly-supervised learning \citep{ren2023proposal,rizve2023pivotal} mitigates this challenge by using approximate or high-level labels, allowing models to generalize without requiring every frame to be manually annotated. As a result, this approach offers a scalable pathway for handling diverse, real-world HOI data, where precise and exhaustive annotations may be neither feasible nor reliable.

\section*{Acknowledgement}
{
This research is supported in part by the EPSRC NortHFutures project (ref: EP/X031012/1).
}

\bibliographystyle{elsarticle-harv}
\bibliography{main.bib}

\end{document}